\documentclass[runningheads]{llncs}

\usepackage{booktabs,siunitx}
\usepackage{adjustbox}

\usepackage{epstopdf}
\AppendGraphicsExtensions{.tif}

\usepackage{enumerate}
\usepackage{multirow}
\usepackage{float} 
\usepackage{array}
\newcommand{\PreserveBackslash}[1]{\let\temp=\\#1\let\\=\temp}
\newcolumntype{C}[1]{>{\PreserveBackslash\centering}p{#1}}

\usepackage{amsmath}
\date{ }

\usepackage{graphicx}
  
\usepackage[dvipsnames]{xcolor}
\begin{document}
%


\title{Fall detection using multimodal data}
%
%
 \author{Thao V. Ha \inst{1,2} \and
 Hoang Nguyen\inst{1,2} \and
 Son T. Huynh \inst{1,2,3} \and Trung T. Nguyen  \inst{4} \and Binh T. Nguyen  \inst{1,2,3}
 \authorrunning{Thao et al.}
 %
 \institute{University of Science, Ho Chi Minh City, Vietnam 
 \and
 Vietnam National University in Ho Chi Minh City, Vietnam
 \and 
 AISIA Research Lab, Ho Chi Minh City, Vietnam 
 \and Hong Bang International University\\
 \email{ngtbinh@hcmus.edu.vn}}}
\maketitle              
\begin{abstract}
In recent years, the occurrence of falls has increased and has had detrimental effects on older adults. Therefore, various machine learning approaches and datasets have been introduced to construct an efficient fall detection algorithm for the social community. This paper studies the fall detection problem based on a large public dataset, namely the UP-Fall Detection Dataset. This dataset was collected from a dozen of volunteers using different sensors and two cameras. We propose several techniques to obtain valuable features from these sensors and cameras and then construct suitable models for the main problem. The experimental results show that our proposed methods can bypass the state-of-the-art methods on this dataset in terms of accuracy, precision, recall, and F1-score. 

\end{abstract}
\section{Introduction}
\label{section:Introduction}
Falling is one of the most common dangers that the elderly usually face during their daily lives, and the potential of death after falling might increase if they live alone.  As reported by the Center for Diseases and Controls (CDC)\footnote{\label{note1}\url{https://www.cdc.gov/homeandrecreationalsafety/falls/adultfalls.html}}, the percentage of deaths after falling in the U.S went up 30\% from 2007 to 2016 for older adults. In case we do not find an appropriate way to stop these rates from continuing to grow, there may be approximately seven deaths per hour by 2030. Among persons above 65 years of age or older, more than one-third of them fall each year, and remarkably, in half of such cases, the falls are recurrent \cite{al2011falls}. The corresponding risk may double or triple with the occurrence of cognitive impairment or history of previous falls \cite{tinetti1988risk}.
Typically, there are various costly consequences that fall incidents lead to, including:
\begin{enumerate}
    \item Causing serious injuries for the elderly such as broken bones e.g. wrist, arm, ankle, and hip fractures.
    \item Causing head injuries for people who are who are taking certain medicines could make their situations worse. Furthermore, when fall incident result in damage to an elderly person’s head, the people need to go to the hospital right away to inspect for any brain injuries.
    \item Causing many people the fear of falling and making them less active. As a result, they become weaker and have a higher percentage of getting the same incident again.
\end{enumerate}

Understanding the fearful outcomes that falling leads to, developing a fall detection system is essential than ever before. In addition, when an incident occurs, the time that the elderly remain to lie on the floor after the fall is one of the critical factors for determining the severity of the fall \cite{igual2013challenges}. Timely detection of falls can quickly help older people to receive immediate assistance by caregivers and then reduce the adverse consequences from the incident \cite{bagala2012evaluation}. 
Consequently, a robust fall detection system to monitor the fall and provide alerts or notifications is necessary to lighten the burden of caregivers and resource-strained health care systems \cite{xu2018new}.

This paper aims to investigate the falling detection problem based on a public dataset, namely the UP-Fall Detection dataset, provided by Martinez and colleagues \cite{martinez2019up}. This dataset contains sensor data and images collected by various devices and sensors, including wearable sensors, ambient sensors, and vision devices, from different healthy young volunteers. They performed six daily activities and simulated five different falls, with three attempts for each activity. The wearable sensors include an accelerometer, gyroscope, and ambient light sensors. On the other hand, they used one electroencephalograph (EEG) headset, six infrared sensors, and two cameras to acquire data. Furthermore, we present an improved method for the fall detection problem in this dataset and compare the proposed approach with previous techniques. The experimental results show that our method could bypass the state-of-the-art techniques and obtain better accuracy, precision, recall, and F1-score.

\section{Related Work}
There have been recent works related to the research of building fall detection systems. For example, Vallabh et al. \cite{vallabh2016fall} introduced their fall detection system using different classifiers, which are: Naïve Bayes, K-nearest neighbor, neural network, and support vector machine. Furthermore, they measured the corresponding performance of these methods based on two well-known datasets (FDD and URFD). In the experiments, Support Vector Machine achieved the best performance with 93.96\% accuracy.

Delgado and colleagues \cite{delgado2020cross} presented a new deep learning-based approach for the fall detection problem using four datasets recorded under different conditions. They utilized sensor data and subject information (accelerometer device, sampling rate, sequence length, age of the subjects, etc.) for feature extraction and obtained more than 98\% of accuracy in these datasets. Furthermore, the proposed platform could get a low false positive (less than 1.6\% on average) and handle simultaneously two tasks: fall detection and subject identification. 

Tsai et al. \cite{tsai2019implementation} presented a fall detection system by combining both traditional and deep neural methods. First, for extracting relevant features for the main problem, they initialized a skeleton information extraction algorithm that could transform depth information into skeleton information and extract the important joints related to fall activity. Then, they pulled seven highlight feature points and employed deep convolution neural networks to implement the fall detection algorithm based on the approach. As a result, they could obtain high accuracy on a popular dataset NTU RGB+D with 99.2\% accuracy. One can find more details at \cite{sadreazami2019tl,keskes2021vision}.

\section{Methodology}

This section introduces our approach to the fall detection problem. First, we describe the feature extraction step with sensor and camera data and then present various models for the fall detection problem based on features extracted. We also provide the list of performance metrics used in our experiments.

\subsection{ UP-Fall Detection dataset}

All volunteers set up different devices to collect the UP-Fall Detection dataset, including wearables, context-aware sensors, and cameras. They collected these multimodal data at the same time. During the data collection process, these volunteers stayed in a controlled laboratory room, having the same light intensity, and the context-aware and cameras remained in the same position. 

\begin{figure}[!ht]
    \centering
    \includegraphics[width=0.99\textwidth]{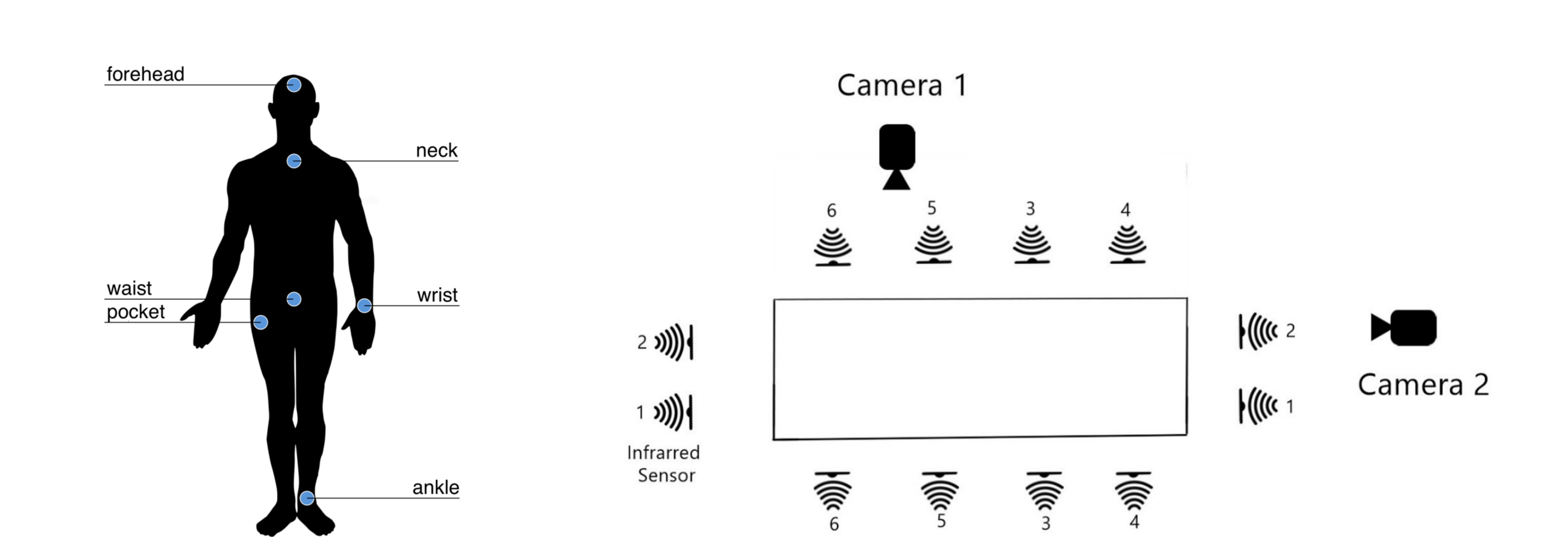}
    \caption{Location of different sensors and camera devices in the  UP-Fall Detection dataset \cite{martinez2019up}}
    \label{fig: losacd}
\end{figure}

Five Mbientlab MetaSensor wearable sensors were put in the five different places (the left wrist, below the neck, in the right trouser pocket, in the middle of the waist (in the belt), and at the left ankle) to collect raw data (the 3-axis accelerometer, the 3-axis gyroscope, and the ambient light value). In addition, each volunteer used one electroencephalograph (EEG) NeuroSky MindWave headset to measure the associated EED signals from the head. Six other infrared sensors were placed as a grid 0.40 m above the room floor to track all changes in interruption of the optical devices. In addition, the authors installed two Microsoft LifeCam Cinema cameras at 1.82 meters above the floor for two different views: lateral view and frontal view (as depicted in Figure \ref{fig: losacd}). One can find further details at \cite{martinez2019up}.

\begin{table}[!ht]
\caption{Activities duration in the  UP-Fall Detection dataset \cite{martinez2019up}}
\label{tab:ada}
\centering
\begin{tabular}{|c|c|c|}
    \hline
    \textbf{Activity ID} & \textbf{Description} & \textbf{Duration (s)} \\
    \hline
  1 & Falling forward using hands & 10 \\
  2 & Falling forward using knees & 10\\
  3 & Falling backwards & 10\\
  4 & Falling sideward & 10 \\
  5 & Falling sitting in empty chair & 10 \\
  6 & Walking & 60 \\
  7 & Standing & 60 \\
  8 & Sitting  & 60 \\
  9 & Picking up an object & 10 \\
  10 & Jumping & 30 \\
  11 & Laying & 60 \\
    \hline
\end{tabular}
\end{table}
\hfill \\

\subsection{Data Processing}

Related to the sensor data, we dropped all duplicate records and removed rows having missing values. Finally, to combine the sensor data with all images extracted from a camera, we carefully checked the timestamp information from the sensor data and selected the most relevant mapping to associated images. As a result, the total number of samples for sensor data is 258,113 with 28 different attributes and one label. 
When extracting useful features for sensor data, we applied the standardization technique for the sensor dataset by normalizing the mean of attributes to zero and the corresponding standard deviation to one. 

We ensured all images extracted from Camera 1 and Camera 2 could have the same size and sorting order for the camera data by removing redundant photos of both cameras. We also carefully reviewed the timestamp of both images and sensor data for the most relevant mapping.
We scaled each image by dividing each pixel's value to 255 to guarantee those entire photos' pixels were in the range [0,1].

\subsection{Feature extraction and modeling}
As described in the previous section, two data sources are collected in the UP-Fall Detection dataset, sensor and camera data. Therefore, we employ different feature extraction steps for these data after doing necessary data processing.

\subsubsection{Sensor data}

\begin{figure}[!ht]
    \centering
    \includegraphics[width=0.81\textwidth]{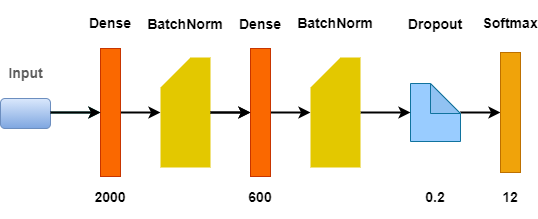}
    \caption{Our proposed neural network for the fall detection algorithm using Sensor data.}
    \label{fig:sensor_model}
\end{figure}

With a given list of 28 attributes from the sensor data, we present the following neural network only using sensor features for the fall detection algorithm: one fully connected layer of 2000 units with Relu activation function, one batch normalization layer, one fully connected layer of 600 units using the Relu activation function, another batch normalization layer, a dropout layer of the rate 0.2, and the final Softmax layer for the output size as 12. One can see more detailed in Figure \ref{fig:sensor_model}. 

Besides using the proposed neural network above, we also consider two other techniques to create a suitable fall detection algorithm using only sensor data: XGBoost and CatBoost.

XGBoost is an optimized gradient tree boosting system that enables the design of decision trees
in a sequential form \cite{zhao2018assessing}. Moreover, this algorithm can compute relevant calculations relatively
faster in all computing environments. As a result, XGBoost is widely used for its performance in
modeling newer attributes and classification of labels.\cite{bhattacharya2020novel}.

At the same time, CatBoost is a strong gradient boosting machine learning technique that achieves
state-of-the-art results in various practical tasks. Despite the original aim of designing this algorithm being to deal with categorical features, it is still plausible to run CatBoost over a dataset with continuous features \cite{al2021feature}. We will show the corresponding results in our experiments.

\begin{table}[!ht]
\caption{Parameters of 2 ML models} 
\centering 
\begin{tabular}{ c|cc } 
\hline \hline
\textbf{Models} & \textbf{Parameters}  \\
\hline
\multirow{5}{15em}{   \hspace{1.5cm}       XGBoost} 
& objective="multi:softprob", \\
& learning rate = 0.5, \\ 
& random state = 42, \\ 
& use label encoder = False, \\ 
& \# of estimators = 100 \\
\hline
\multirow{3}{15em}{  \hspace{1.5cm}  CatBoost} 
 &  \# of estimators = 500, \\
 &  random seed = 42, \\
 &  learning rate = 0.25, \\
 &  max depth = 12 \\
\end{tabular}
\label{tab:Parameter}
\end{table}

\subsubsection{Camera data}


We employed convolutional neural networks (CNNs) to extract features from camera data. It is worth noting that CNN has been performing outstanding results for understanding contents presented in an image better and achieving state-of-the-art results in different applications, including image recognition, segmentation, detection, and retrieval \cite{sharma2018analysis}. 

It is worth noting that there are two cameras installed in the  UP-Fall Detection dataset: Camera 1 and Camera 2. As a result, we consider three different cases for constructing an appropriate fall detection model for the main problem.

\begin{figure}[!ht]
    \centering
    \includegraphics[width=0.81\textwidth]{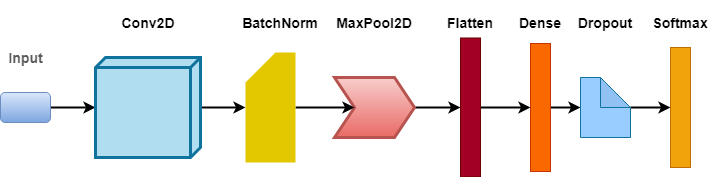}
        \caption{Our proposed CNN for constructing a suitable fall detection model only using one camera (Camera 1 or Camera 2).}
\label{fig:CNN_only_camera}
\end{figure}

For only using Camera 1 or Camera 2, we selected the input size of images collected as (32,32). Then, we pushed the input data to the same CNN architecture having the following layers: a two-dimensional convolutional layer with 16 filters of size (3,3), one batch normalization layer, one Max-pooling layer of size (2,2), one Flatten layer, a fully-connected layer of 200 units, one Dropout layer of the rate 0.2, and the final Softmax layer with 12-dimensional output. We depicted this CNN in Figure \ref{fig:CNN_only_camera}.

\begin{figure}[!ht]
    \centering
    \includegraphics[width=0.81\textwidth]{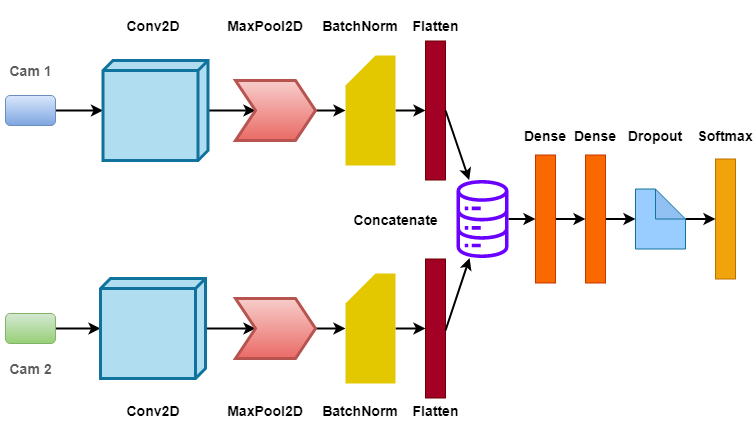}
    \caption{Concatenated CNN model for Cam 1 + Cam 2}
    \label{fig: CCNN1}
\end{figure}

Finally, we study the remaining case when combining images collected from Camera 1 and Camera 2. Typically, we extracted features from each camera for given input data. As a result, the input data from Camera 1 and Camera 2 shifted through the same CNN architecture: one two-dimensional convolutional later with the number of filters as 15 and the kernel size as (3,3), one Max pooling layer with the pool size as (2,2), one batch normalization later, and one flattened layer. After this step, all two features extracted from Camera 1 and Camera 2 could be concatenated and then go through two consecutive fully-connected layers with the corresponding number of units as 400 and 200 using the Relu activation function. We used another dropout after that for regularization, and this could help us reduce the percentage of overfitting problems during the training step. Subsequently, we put the computed vector into the final layer using the Softmax activation function to obtain the 12-dimensional output. One can see more details in Figure \ref{fig: CCNN1}.

\subsubsection{Fusion data}

\begin{figure}[!ht]
    \centering
    \includegraphics[width=0.81\textwidth]{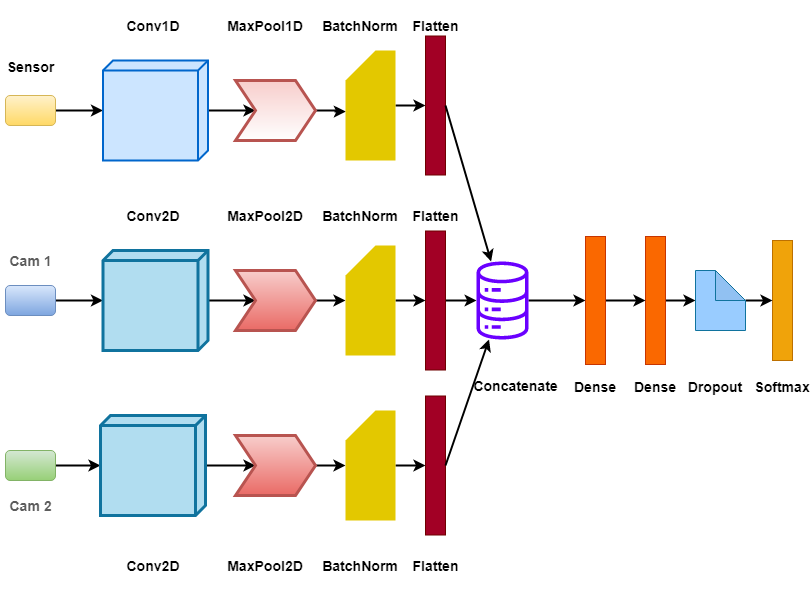}
    \caption{Our proposed deep neural network using all sensor data and two cameras.}
    \label{fig: CCNN}
\end{figure} 
   
For fusing multimodal data from both sensors and cameras, we designed the following neural network architecture. First, two input data from two cameras were passed through the same CNN architectures as mentioned above. Then, on the other hand, sensor data are passed through one convolutional 1D layer with ten filters and the size of the kernel as three, and the Relu activation function. Next, these computed vectors from sensors and two cameras continue passing through one max pooling 1D layer with the pool size as two, one batch normalization layer, and a flattened layer. Subsequently, these three flattened layers are concatenated as a final feature vector before going through two fully connected layers with units 600 and 1200. Next, the dropout layer with a rate of 0.2 is added for regularization, and the final result can be computed via a Softmax layer at the end for classifying fall detection. One can find more information related to this architecture in Figure \ref{fig: CCNN}.

\subsection{Performance metrics}
For comparing the performance of different approaches, we use the following metrics in our experiments: accuracy, precision, recall, and F1 scores.

\section{Experiments}
This paper runs all experiments on a computer with Intel(R) Core (TM) i7-7700K 4 CPUs running at 4.2GHz with 16GB of RAM and 48GB of virtual memory. During the processing step, with the help of some essential libraries: \textbf{numpy}, \textbf{pandas}, \textbf{cv2}, we can work through the operation easily. In the modeling procedure, the \textbf{scikit-learn} package provides a powerful tool for us to run some algorithms like XGBoost and CatBoost. On the other hand, the \textbf{Tensorflow} and \textbf{Keras} libraries are crucial tools to train deep learning models. In the end, the package \textbf{ModelCheckpoint} is utilized to save deep learning models and \textbf{joblib} for \textbf{scikit-learn} models.

\subsection{Data Collection} 
\label{subsec : dd}

We used the UP-Fall Detection dataset for all experiments. This dataset was published by Martinez et al. \cite{martinez2019up} at the following link\footnote{\url{https://sites.google.com/up.edu.mx/har-up}}.
There are two types of datasets in this link: Consolidated Dataset and Feature Dataset. We decided to use the Consolidated Dataset because it is the core dataset to make further extractions more accessible. 
We concatenate all the CSV files together for the sake of easiness in the training process. After combining all the files, we get a CSV file with 294,678 samples and 45 features.

All images are converted to gray-scale images and resized to the shape of (32,32) by the following equation:
\begin{equation*}
    Gray = 0.299 * Red + 0.587 * Green + 0.114 * Blue
\end{equation*}

\subsection{Previous methods}

In the previous work proposed by Martinez et al. \cite{martinez2019up}, the authors did not give the information about a random seed for the dataset to reproduce the result with their techniques. As a result, we decided to split the dataset into training, test, and validation sets with the ratio of $60/20/20$ and run experiments with our proposed models for both sensor and camera data. Likewise, the models in the article \cite{martinez2019up} are implemented again to compare the performance with our proposed model. From that, we can find a better fall detection system. 

Related to the sensor data, Martinez and colleagues used Random Forest\cite{rf}, Support Vector Machines \cite{svm}, Multi-Layer Perceptron \cite{mlp}, and K-Nearest Neighbors \cite{knn}. One can find more information about in their paper \cite{martinez2019up} as well as their hyperparameter configuration in table \ref{tab:Parameter}.

\begin{figure}[!ht]
    \centering
    \includegraphics[width = 0.81\textwidth]{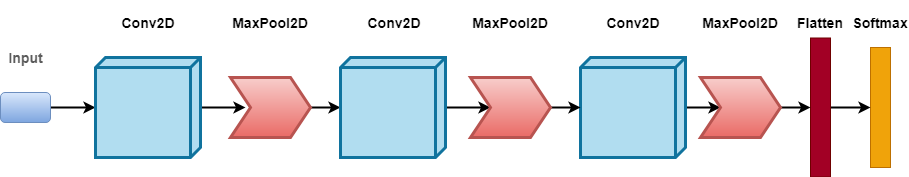}
    \caption{The CNN model adapted from \cite{martinez2019up}}
    \label{fig:r1}
    \end{figure}

For using camera data, they implemented one model using a Convolutional layer with eight filters size $3\times3$, one ReLu activation function, and then one Max Pooling layer of size $2\times2$. These consecutive layers repeated twice in that architecture with minor changes in filter sizes of the Convolution layer: 16 in the second and 32 in the third time. Finally, this model ended with one flatten layer and a Softmax layer with size 12 for obtaining the final prediction. It is worth noting that they trained this system using the stochastic gradient descent algorithm with an initial learning rate of 0.001, regularization coefficient 0.004, a maximum number of epochs 5, and a mini-batch size of 100. This architecture can be shown in Figure \ref{fig:r1}.
\subsection{Experimental results}
In experiments, we analyze the performance of three different methods: using the sensor dataset (S), only using two cameras that are Camera 1 (C1) and Camera 2 (C2), a combination of two cameras (C1+C2), and a compound of these features (S+C1+C2). Then we measure all results using four metrics, including Accuracy, Precision, Recall, and F1-Score. One can see more details in our experimental results in Tables \ref{tab: poo}, \ref{tab:poo1}, and \ref{tab:cm}.

First, we compare models and techniques on the sensor dataset without using any Camera information. The experiment shows that three of our algorithms, including XGBoost, CatBoost, and Multi-Layers Perceptron (MLP), can achieve exhilarating results in terms of Accuracy, Precision, Recall and F1-Score greater than $99\%$. Interestingly, our methods could bypass all techniques that Martinez et al. \cite{martinez2019up} which consists of Random Forest, Support Vector Machines, Multi-Layer Perceptron, and K-Nearest Neighbors. Significantly, in our methods, we increase the number of layers and units in the MLP model. As a result, the critical metric, F1-Score, can reach $99.03\%$ with the modified MLP, while the best result of the previous work barely gained $97.28\%$.
\begin{table}[!ht]
\caption{Parameters of ML models in \cite{martinez2019up}} 
\label{tab:PFML} 
\centering 
\resizebox{12cm}{!}{
\begin{tabular}{C{7cm}|C{7cm}} 
\hline 
\textbf{Models} & \textbf{Parameters}  \\
\hline
\multirow{4}{15em}{        Random Forest} 
& estimators = 10, \\
& min. samples split = 2, \\ 
& min. samples leaf = 1, \\ 
& bootstrap = true \\ 
\hline
\multirow{5}{15em}{   Support Vector Machines} 
 &  c = 1.0, \\
 &  kernel = radical basis function, \\
 &  kernel coefficient = 1/features, \\
 &  shrinking = true,\\
 & tolerance = 0.001 \\
 \hline
\multirow{3}{15em}{  Multi-Layer Perceptron} 
 &  hidden layer size = 100, \\
 &  activation function = ReLU, \\
 &  solver = stochastic gradient, \\
 &  penalty parameter = 0.0001, \\
 &  batch size = min(200,samples)
, \\
 &  initial learning rate = 0.001, \\
 &  shuffle = true, \\
 &  tolerance = 0.0001
, \\
 &  exponential decay (first moment) = 0.9, \\
 &  exponential decay (second moment) = 0.999, \\
 &  regularization coefficient =  0.000000001, \\
 &  max. epochs = 10 \\
 \hline
\multirow{3}{15em}{   k-Nearest Neighbors
} 
 &  neighbors = 5
, \\
 &  leaf size = 30, \\
 &  metric = Euclidean \\
\end{tabular}}
\label{tab:PPer}
\end{table}
For the approach using Camera information, we modify the CNN model, proposed by Martinez et al. \cite{martinez2019up} by eliminating two Conv2D layers and adding BatchNormalization and Dropout. Our proposed method reduces the time training and increases the performance in Accuracy, Precision, Recall higher than $99.1\%$ in Camera 1 and $99.3\%$ in Camera 2. Furthermore, the F1-Score of Camera 1 and Camera 2 can achieve $99.16\%$ and $99.40\%$ respectively. In comparison, model CNN of the previous author group \cite{martinez2019up} has the results of F1-Score value in Camera 1 and Camera 2 are $76.69\%$ and $86.96\%$ in that order. From this result, using multiple Convolutional layers as the previous work \cite{martinez2019up} in a consecutive sequence does not always give good results. One possible reason is that more information can be lost for each time passing through a Convolutional layer. In this case, using a convolutional layer is enough to extract information from the input image. In addition, BatchNormalization layers make the model easier to converge and overfit in the training set. Finally, adding Dropout classes helps the model avoid overfitting, thereby providing high performance on the test set. 

Interestingly, using both the information of the sensor and two Cameras can help improve our approach's performance. In the case of the concatenation models in Table \ref{tab:cm}, the results are far better in all metrics that we mentioned above. At first, we try to combine Camera 1 and Camera 2 to train with this model, and it gets $99.46\%$ in F1-Score compared to the best model we acquired with CNN, which is $99.40\%$ F1-Score. Furthermore, it is worth noting that a training model with the combination of sensors and two Cameras can outperform using each feature. The corresponding Accuracy, Precision, Recall, and F1-Score are $99.56\%$, $99.56\%$, $99.56\%$, $99.55\%$, which is most dominant than the using Sensor, one Camera, and the combination of two Cameras on each metric.

\begin{table}[!ht]
\caption{Performance of our proposed models}
\label{tab: poo}
\centering
\begin{tabular}{|c|c|c|c|c|c|}
    \hline
    \textbf{Data} & \textbf{Model} & \textbf{Accuracy} & \textbf{Precison} & \textbf{Recall} & \textbf{F1-Score} \\
    \hline
    \multirow{3}{1em}{\textbf{S}}
    &XGBoost & 99.21 & 99.19 & 99.21 & 99.20 \\
    &Catboost & 99.05 & 99.02 & 99.05 & 99.02\\
    &MLP & 99.04 & 99.05 & 99.03 & 99.03\\
    \hline
    \textbf{C1}
    &CNN & 99.17 & 99.24 & 99.12 & 99.16 \\
    \hline
    \textbf{C2}
    &CNN & 99.39 & 99.40 & 99.39 & 99.40 \\
    \hline
\end{tabular}
\end{table}

\begin{table}[!ht]
\centering 
\caption{Performance of models of Martinez et al. \cite{martinez2019up}}
\label{tab:poo1}
\begin{tabular}{|c|c|c|c|c|c|}
    \hline 
     \textbf{Data} & \textbf{Model} & \textbf{Accuracy} & \textbf{Precison} & \textbf{Recall} & \textbf{F1-Score} \\
    \hline
    \multirow{4}{1em}{\textbf{S}} 
    & RF & 97.46 & 97.29 & 98.46 & 97.28 \\
    & SVM & 96.96 & 96.82 & 96.96 & 96.61\\
    & KNN & 97.24 & 97.07 & 97.24 & 97.05\\
    & MLP & 90.21 & 88.36 & 90.21 & 88.43\\
    \hline
    \textbf{C1}
    &CNN & 78.92 & 84.80 & 70.97 & 76.69 \\
    \hline
    \textbf{C2}
    &CNN & 88.24 & 90.32 & 86.13 & 86.96 \\
    \hline
      \end{tabular} 
\end{table}

\begin{table}[!ht]
\centering 
\caption{The performance when combining different features: C1+C2 and S+C1+C2.}
\label{tab:cm}

\makebox[\textwidth]{
\begin{tabular}{|c|c|c|c|c|c|}
\hline 
 \textbf{Data} & \textbf{Model} & \textbf{Accuracy} & \textbf{Precison} & \textbf{Recall} & \textbf{F1-Score} \\
\hline
\textbf{C1+C2}
&Combination & 99.46 & 99.47 & 99.46 & 99.46 \\
\hline
\textbf{S+C1+C2}
&Combination & 99.56 & 99.56  & 99.56  & 99.55  \\
\hline

\end{tabular} } 
\end{table}
\section{Conclusion and Future Works}
We have proposed a new approach for the fall detection problem by using the concatenation model. With this methodology, we can combine different kinds of data and find a new path to improve the model's performance, which is critical in developing a fall detection system. The experiments show that fusing both sensor and camera data can improve performance for the fall detection algorithm in the UP-Fall Detection dataset.

In the future, we aim to focus more on feature extractions work on this dataset to have a deeper understanding of falling. In addition, we will also apply other recent techniques in this data to improve the performance of the fall detection problem.
\medskip

\bibliographystyle{IEEEtran}
\bibliography{references}

\end{document}